# Improving the Quality of MT Output using Novel Name Entity Translation Scheme

Deepti Bhalla
Department of Computer Science
Banasthali University
Rajasthan, India
deeptibhalla0600@gmail.com

Nisheeth Joshi
Department of Computer Science
Banasthali University
Rajasthan, India
nisheeth.joshi@rediffmail.com

Iti Mathur
Department of Computer Science
Banasthali University
Rajasthan, India
mathur_iti@rediffmail.com

*Abstract*— This paper presents a novel approach to machine translation by combining the state of art name entity translation scheme. Improper translation of name entities lapse the quality of machine translated output. In this work, name entities are transliterated by using statistical rule based approach. This paper describes the translation and transliteration of name entities from English to Punjabi. We have experimented on four types of name entities which are: Proper names, Location names, Organization names and miscellaneous. Various rules for the purpose of syllabification have been constructed. Transliteration of name entities is accomplished with the help of Probability calculation. N-Gram probabilities for the extracted syllables have been calculated using statistical machine translation toolkit MOSES.

*Keywords— machine translation; machine transliteration; name entity translation; n-gram probability; syllabification*

## I. INTRODUCTION

We live in a world which constitutes various languages that reflect its linguistic diversity. Machine translation is the field which comes under the area of computational linguistics and is also one of the emanating research areas of Natural Language Processing. Machine translation is the process in which the syntactic/semantic structure of our source language is converted to the syntactic/semantic structure of target language. For the purpose of this translation, it is very crucial to obtain the rigorous translation of named entities. Name entity recognition has many applications in the cross-lingual information extraction such as information retrieval, text based question answering etc. For the recognition of name entities, we have used Stanford NER tool [1] which recognizes 4 types of name entities. We have mainly emphasized on statistical machine translation to accomplish our work. For translating our text from language of its generation to the language of its aspiration we have used corpus based approach.

The name entities of type location and organization are passed to the translation module and the name entities of type person and miscellaneous are passed to the transliteration module which constitute the syllabification process before probability matching. Machine transliteration deals with conversion of the characters in the input string from source language to the target language while preserving its phonological structure. In this paper transliteration is done with the help of syllable extraction. We have used phonetic based statistical approach for transliteration.

The remainder of this paper is organized as follows: In section II we described related work. Section III explains our methodology and experimental set up. Section IV consists of evaluation and results. Finally in section V we conclude the paper.

## II. RELATED WORK

A lot of research work has been carried out in the area of machine translation and machine transliteration. Singh et al. [2] presented statistical analysis of syllables and have shown that how this statistical analysis will be helpful in selection of syllables for the speech database. They have analyzed the syllables over a large Punjabi corpus having more than 104 million words. Gupta et al. [3] implemented a condition based named entity recognition algorithm. For the implementation of NER, they have developed various resources in Punjabi, like a list of prefix names, a list of suffix names, a list of proper names, middle names and last names. For condition based system they have done in-depth analysis of output of over 50 Punjabi news documents. Kamal Deep and Goyal [4] have addressed the problem of transliterating text from Punjabi to English language by using the rule based approach. To model the transliteration problem the proposed scheme uses various approaches such as Grapheme based approaches ($\Psi_G$), phoneme-based approaches ($\Psi_P$) and hybrid approaches ($\Psi_H$). Their technique has demonstrated the transliteration from Punjabi to English for common names. This technique has achieved accuracy of 93.22%. Babych and Hartley [5] reported the results of an experiment on automatic NE recognition from Machine Translations produced by five different MT systems. They have used outputs from name entity recognition module of sheffield's GATE information extraction system. They have shown the result which indicates that combining IE technology with Machine translation has a great potential for improving the state of art in output quality. Yasar AL-Onaizan and Knight [6] presented a novel algorithm for translating Name Entity phrases using easily obtainable monolingual and bilingual resources and the results obtained are compared with the results obtained from human translations and a commercial system for the same task. They described a system for Arabic-English Name Entity Translation and after comparison with the human translator they achieved accuracy of 84%. Hassan et al. [7] evaluated the quality of the extracted translation pairs by showing that it improves the performance of a named entity translation system. They used this approach to build a large dictionary of Arabic/English named entity translation pairs. Alek et al. [8] presented a approach to improve machine translation of named entities by using Wikipedia. In this the author addressed various problems by using Wikipedia to translate NEs and presented them, already translated, to the MT system. They experimented with English to Czech translation. Kamal deep et al. [9] proposed hybrid (statistical +rules) approach for Punjabi to English transliteration system of person names, in that for a person name written in Punjabi (Gurumukhi Script), their system produces its English (Roman Script) transliteration. The Authors used letter to letter mapping as baseline and tried to find out the improvements by using statistical methods. They achieved accuracy of 95.23%. Batra et al. [10] proposed Rule Based Machine Translation of Noun Phrases from Punjabi to English. They have presented automatic translation of noun phrases from Punjabi to English using transfer approach. Their system includes analysis, translation and synthesis component. Their system achieved accuracy of 85%. Gupta et al. [11] have shown the previous work done in English and other European languages. A survey is given on the work done in Indian Languages i.e. Telugu, Hindi, Bengali, Oriya and Urdu. They have listed and categorized the features that are used in recognition of NE and also provided an overview of the evaluation methods that are used in calculating the NER accuracy. Musa et al. [12] proposed Syllabification Algorithm based on Syllable Rules Matching for Malay Language. They evaluated there method using Bernama, Kamus Dewan and Overlap data collection. In this paper, authors presented a new syllabification algorithm for Malay language. Joshi and Mathur [13] proposed the use of phonetic mapping based English-Hindi transliteration system which created a mapping table and a set of rules for transliteration of text. Joshi et al. [14] also proposed a predictive approach of for English-Hindi transliteration where the authors provided a suggestive list of possible text that the user entered. They looked at the partial text and tried to provide possible complete list as the suggestive list that the user could accept or provide their own input text. The use of transliteration has been proposed by many researchers for natural language processing and information retrieval applications. Both these approaches were used in approach proposed by Bhalla et al. [15] who used them for transliterating person and location names.

### III. METHODOLOGY AND EXPERIMENTAL SETUP

The objective of our work is to translate the name entities of type location and organization and to transliterate the proper names and miscellaneous entities using syllable extraction. The system architecture given in figure 1 follows various steps through which our input in source language has to be passed in order to convert it in to output text in target language. The overall process is carried out as follows-

#### A. Name Entity Recognition

This is the first layer of our experimental setup. For the conversion of our input text in English to Punjabi we have used 4-class name entity recognition tool. This tool gives us name entities in English with their specific Tag sets. Name entities recognized by this tool include person names, location names, organization & miscellaneous entities. For example-**Mina is going to Hyderabad**

The recognized name entities are-
Mina/**PERSON**
Hyderabad/**LOCATION**

#### B. Pre-processing

This is the second step which gets the name entities with their Tag sets in English. This step simplifies the input text by applying some pre-processing steps. This step separates all unwanted input text such as Tag sets, commas, hyphens etc, only the name entities are passed to the translation/transliteration module for further processing. For example- **Priyanka is going to Delhi University**
We get the following output after the recognition process-
Priyanka/**PERSON**
Delhi University/**ORGANIZATION**

The pre-processing step gives us the following output-
Priyanka
Delhi University

#### C. Translation

In this step name entities will be passed to the translation module. Before passing the name entity to the translation module it is analyzed to check whether it corresponds to the category of location and organization. After the analysis if the name entity is of type location and organization, it is passed to the translation module otherwise it is passed to the syllabification module for further transliteration. Translation module is linked to the knowledge base.
If there is a name entity of type organization then its usual translation for an MT system, which do not have an NER will be:
English- Indian Institute of Technology/**ORGANIZATION**

Punjabi- ਭਾਰਤੀ ਸੰਸਥਾਨ ਦਾ ਤਕਨੀਕੀ (*Bharti sansthan da takniki*)

The system which has the NER module will provide the translation as:

ਭਾਰਤੀ ਤਕਨੀਕੀ ਸੰਸਥਾਨ (*Bharti Takniki Sansthan*)

This translation is accomplished with the help of a knowledge base.

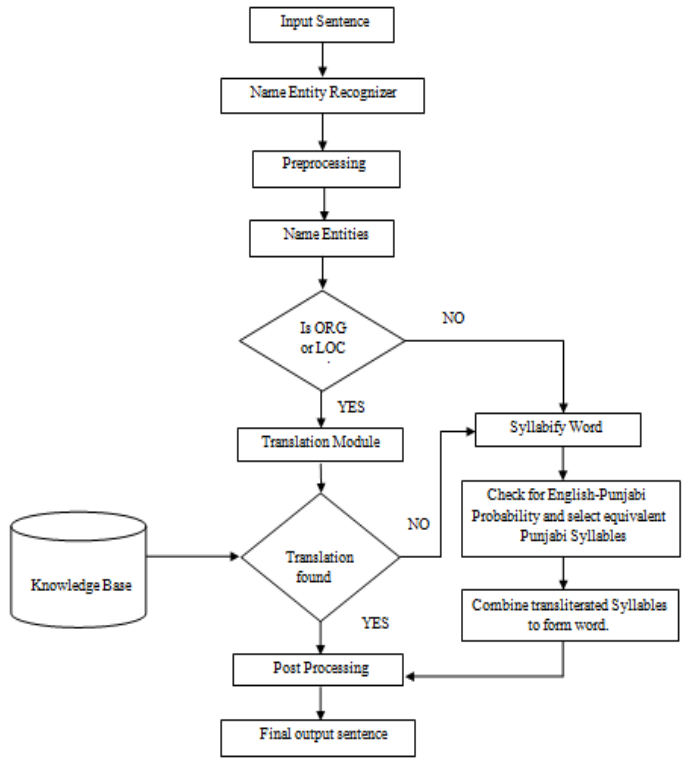

Fig. 1 System Architecture

### D. Syllabification

Syllabification basically is to extract the syllables from a word. We have constructed the rules for transliteration using the syllabification approach. The syllable is a combination of vowel & consonant pairs such as V, VC, CV, VCC, CVC, CCVC and CVCC. Almost all the languages have VC, CV or CVC structures so we have used vowel and consonants as the basic phonological unit. The statistical approach used for syllabification is based on phonetic matching.

**Algorithm for syllable extraction**

1. Define English letter set as E, Vowel set as V= {a,e,i,o,u} and Consonant set as C.
2. The continuous Vowels should be considered as separate Vowel set.
3. In case of Vowel consideration there are some special cases such as ie, io which are considered as individual syllables. For example ubiety will be syllabified as u/bi/e /ty.
4. In case of continuous Vowels they can be cut in to several independent Consonants or as a separate set.
5. There are some continuous Consonants which are pronounced together such as sh, gh, ty, ny. For example ability is syllabified to a/bi/li/ty.
6. There are some other syllables also which combinations of various Vowels and Consonants are and they are always pronounced together such as tion, sion, ment. These syllables are always considered as a single syllable.

Some of the possible rules constructed for syllable extracting are shown in Table I:

TABLE I
POSSIBLE SYLLABLE COMBINATION

| Syllable structure | *Example* | *Syllabified form(English)* | *Syllabified form(Punjabi)* |
|---|---|---|---|
| V | Aya | [a] [ya] | [ਅ] [ਜਾ] |
| CV | Silki | [sil] [ki] | [ਸਿਲ] [ਕੀ] |
| VC | Ashka | [ash] [ka] | [ਆਸ਼] [ਕਾ] |
| CVC | Ridhima | [ri] [dhi] [ma] | [ਰੀ] [ਧੀ] [ਮਾ] |
| CCVC | Orissa | [o] [ri] [ssa] | [ਓ] [ਰੀ] [ਸਾ] |
| CVCC | Abhika | [a] [bhi] [ka] | [ਅ] [ਭੀ] [ਕਾ] |
| VCC | Aya | [a] [ya] | [ਅ] [ਜਾ] |

**V=Vowel, C=Consonant**

### E. Transliteration

We are using the statistical rule based approach for the purpose of transliteration. We have trained our system using the SMT tool MOSES. With the help of GIZA++ [16] we have calculated the translation probabilities on the basis of which we will transliterate our input text. We have redirected the name entities with Tag sets /PERSON and /MISCELLANEOUS directly to the transliteration module. For rest of the name entities if their translations are missing then those name entities are also passed to the transliteration module. After the selection of equivalent Punjabi transliteration on the basis of their probability for all the syllables forming a word, this Punjabi text is passed to the post-processing step and final output is obtained after their combination. Table II shows the test corpus.

TABLE II
TEST CORPUS

|  | English Words |
|---|---|
| Training | 50,000 |
| Testing | 37343 |

Next we performed testing on to our system. Name entities are extracted and transliterated using the parallel corpus of syllabified name entities. We have calculated N-gram probabilities based on relative frequencies for all the extracted syllables. Figure 2 shows the testing architecture.

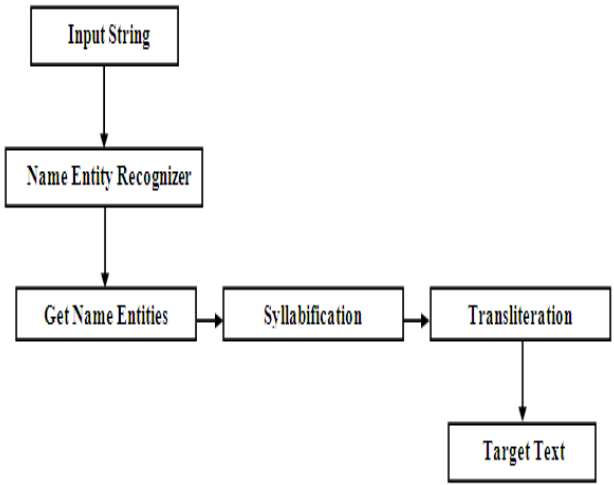

Fig. 2 Testing Architecture

Mohit is going to Haryana with Kunal

Initially with the help of Stanford NER the name entities are extracted which are Mohit/PERSON, Haryana/LOCATION, Kunal/PERSON. The system now checks for the name entities of particular Tag set, if the name entity belongs to PERSON or MISCELLANEOUS Tag set it is directly passed to the syllabification module for transliteration. The syllabification module uses syllable extraction algorithm to divide the words in to syllables.

For the person names **Mohit** and **Kunal** syllabification would be **Mo hit** and **Ku nal** and **Haryana** will be syllabifies as **Har ya na.** Through language model training we have calculated the N-Gram probabilities; these words are now checked against their probability and the transliteration for a syllable is selected according to their probability score. The probability with highest score is selected.

For calculating the N-gram probability we are using the following formula [17].

$$P(w_n|w_{n-1}) = \frac{C(w_{n-1}|w_n)}{C(w_{n-1})} \qquad (1)$$

Where probability of a word $w_n$ given a word $w_{n-1}$ is:

$$P(w_n|w_{n-1}) = \frac{Count(w_{n-1}|w_n)}{Count(w_{n-1})} \qquad (2)$$

For example: For a name Dileep probabilities for every syllable can be as follows:

$Prob(ਦਿ|di) = \frac{C(di,ਦਿ)}{C(di)} = 0.9519231$

$Prob(ਲੀਪ|leep) = \frac{C(leep,ਲੀਪ)}{C(leep)} = 0.5789474$

Final Score= Prob(ਦਿ|di)×Prob(ਲੀਪ|leep) = 0.551113404

After selection of highest probabilities the words are joined together along with their Tag sets and final output is obtained.

## IV. EVALUATION

### A. Evaluation Metrices

#### 1) Accuracy

The main objective of our system is effective translation or transliteration of our recognized name entities. Thus for this we have conducted an accuracy test. We have separately calculated the accuracy for correct translations and transliterations. Table III shows both the Test sets. To measure the quality of our output we have used the following formula-

$$Accuracy(\%) = \frac{correct\ Translations}{Total\ Words} * 100 \qquad (3)$$

$$Accuracy(\%) = \frac{correct\ Transliterations}{Total\ Words} * 100 \qquad (4)$$

We have divided our dataset mainly into two parts. One for the name entities with Tag set PERSON and MISCELLANOUS and another with Tag set ORGANIZATION and LOCATION. Table IV shows the accuracies for both the Test sets.

#### 1) Precision and Recall

From Test set 1 our system correctly transliterated 17693 name entities and from Test set 2 our system correctly translated 14839 name entities. Table V shows the precision and recall factor. Some of the names which are correctly translated/transliterated by our system are shown in Table VI and some which are incorrect are shown in Table VII. Figure 3 shows us the Evaluation results for Test set I and figure 4 shows us evaluation results for Test set II.

TABLE III
TESTING TABLE

| TEST SET 1 | PERSON NAME<br>MISCELLANEOUS | 12548<br>7350 |
|---|---|---|
| TEST SET 2 | LOCATION<br>ORGANIZATION | 9845<br>7600 |

$$\text{Precision} = \frac{\text{Total Correct Translations/Transliteration}}{\text{Total No. of Words}} *100 \quad (5)$$

$$\text{Recall} = \frac{\text{Total correct translation or transliterations by system}}{\text{Total correct translatins/transliterations}} *100 \quad (6)$$

$$F-\text{measure}(\%) = \frac{2.P.R}{(P+R)} *100 \quad (7)$$

TABLE IV
ACCURACIES FOR BOTH TEST SETS

| *TEST SET* | *ACCURACY (%)* |
|---|---|
| **TEST SET 1 (PERSON, MISC)** | 88.91 |
| **TEST SET 2 (LOCATION, ORG)** | 85.06 |

The average accuracy is 86.98%.

TABLE V
PRECISION AND RECALL FACTOR

| *Name Entities* | *Precision(%)* | *Recall(%)* | *F-measure (%)* |
|---|---|---|---|
| **PERSON/MISC (500)** | 87.33 | 80.22 | 83.62 |
| **LOCATION/ORG (500)** | 79.78 | 81.62 | 80.68 |

I. CONCLUSION

In this paper we have constructed name entity translation system based on syllable extraction. If corresponding translation for a name entity is not found then this system transliterates the name entities by matching the probabilities of extracted syllables. Experimental results shows that name entity recognition apparently improves the performance of machine translated output from English to Punjabi. Our system achieved accuracy of 86.98%. In future we will extend our work by taking in to consideration all the possible classes of name entities which are not included in this experiment.

TABLE VI
CORRECT OUTPUT

| **Input Word** | **Correct Output** |
|---|---|
| Harpreet | ਹਰਪ੍ਰੀਤ |
| Delhi | ਦਿੱਲੀ |
| Haryana | ਹਰਿਆਣਾ |
| Mathurawale | ਮਥੁਰਾਵਾਲੇ |

TABLE VII
INCORRECT OUTPUT

| **Input Word** | **Correct Output** | **Our System** |
|---|---|---|
| Arora | ਅਰੋੜਾ | ਅਰੋਰਾ |
| Kaushal | ਕੌਸ਼ਲ | ਕੌਉਸ਼ਲ |
| Pondicherry | ਪੋਂਡੀਚੇਰੀ | ਪੋਂਦੀਚੇਰੀ |
| Sign Of Technology | ਤਕਨੀਕੀ ਚਿਨ੍ਹ | ਤਕਨੀਕੀ ਦਾ ਚਿਨ੍ਹ |

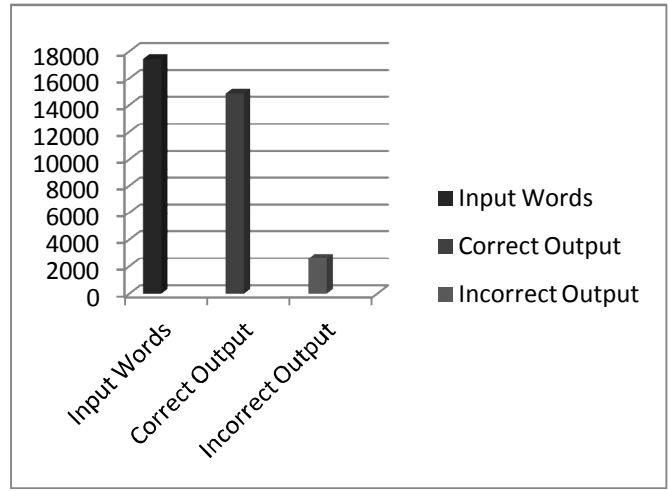

Fig. 3   Evaluation results for Test set I (Transliteration)

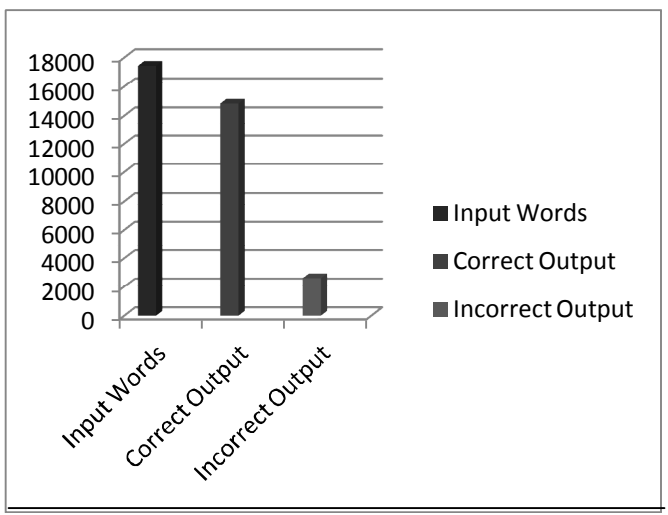

Fig. 4 Evaluation results for Test set II (Translation)

## II. REFERENCES

[1] Jenny Rose Finkel, Trond Grenager, and Christopher Manning. Incorporating Non-local Information into Information Extraction Systems by Gibbs Sampling. *Proceedings of the 43nd Annual Meeting of the Association for Computational Linguistics (ACL 2005),* pp. 363-370, 2005.

[2] Parminder Singh, Gurpreet Singh Lehal, "Corpus Based Statistical Analysis of Punjabi Syllables for Preparation of Punjabi Speech Database". *International Journal of Intelligent Computing Research (IJICR), Volume 1, Issue 3, June 2010.*

[3] Vishal Gupta, Gurpreet Singh Lehal, "Named Entity Recognition for Punjabi Language Text Summarization". *International Journal of Computer Applications (0975 – 8887) Volume 33– No.3, November 2011.*

[4] Kamal Deep, Vishal Goyal, "Development of a Punjabi to English transliteration system". *International Journal of Computer Science and Communication Vol. 2, No. 2,, pp. 521-526, July- December 2011.*

[5] Bogdan Babych, Anthony Hartley, "Improving Machine Translation Quality with Automatic Named Entity Recognition". *Proceedings of the 7th International EAMT workshop on MT and other Language Technology Tools, 1-8, 2003/4/13.*

[6] Yasar AL-Onaizan, Kevin Knight, "Translating Name Entities Using Monolingual and Bilingual Resources". *Proceedings of the 40th Annual Meeting of the Association for Computational Linguistics (ACL), Philadelphia, pp. 400-408, 2002.*

[7] Ahmed Hassan, Haytham Fahmy and Hany Hassan, "Improving Named Entity Translation by Exploiting Comparable and Parallel Corpora" *AMML07, 2007.*

[8] Ond_rej H_alek, Rudolf Rosa, Ale_s Tamchyna, and Ond_rej Bojar proposed a paper on "Named Entities from Wikipedia for Machine Translation", *Proceedings of the Conference on Theory and Practice of Information Technologies, 2011.*

[9] Kamal Deep, Dr.Vishal Goyal, "Hybrid Approach for Punjabi to English Transliteration System". *International Journal of Computer Applications (0975 – 8887) Volume 28– No.1, August 2011.*

[10] Kamaljeet Kaur Batra, G S Lehal, "Rule Based Machine Translation of Noun Phrases from Punjabi to English" in proceedings of *IJCSI International Journal of Computer Science Issues, Vol. 7, Issue 5, September 2010.*

[11] Darvinder kaur, Vishal Gupta "A survey of Named Entity Recognition in English and other Indian Languages" in proceedings of *IJCSI International Journal of Computer Science Issues, Vol. 7, Issue 6, November 2010.*

[12] Musa, Hafiz, Rabith A.kadir, Azreen Azman, M.taufik Abadullah "Syllabification algorithm based on syllable rules matching for Malay language." *Proceedings of the 10th WSEAS international conference on Applied computer and applied computational science.* World Scientific and Engineering Academy and Society (WSEAS) *2011.*

[13] Nisheeth Joshi and Iti Mathur, Input Scheme for Hindi Using Phonetic Mapping. In Proceedings of the National Conference on ICT: Theory, Practice and Applications, 2010.

[14] N. Joshi, I. Mathur and S. Mathur, Frequency Based Predictive Input System for Hindi. In Proceedings of the International Conference and Workshop on Emerging Trends in Technology, ACM, pp 690-693, 2010.

[15] Deepti Bhalla, Nisheeth Joshi and Iti Mathur, Rule Based Transliteration Scheme for English to Punjabi. International Journal of Natural Language Computing, Vol 2, No. 2, pp 67-73, 2013.

[16] To download IRSTLM toolkit http://www.statmt.org

[17] Daniel Jurafsky, James H. Martin Speech and Language processing An Introduction to speech Recognition, natural language processing, and computational linguistics.